\documentclass[11pt]{article}

\usepackage[final]{acl}

\usepackage{times}
\usepackage{latexsym}
\usepackage[T1]{fontenc}
\usepackage[utf8]{inputenc}
\usepackage{microtype}
\usepackage{inconsolata}
\usepackage{graphicx}
\usepackage{booktabs}
\usepackage{amsmath}
\usepackage{amssymb}
\usepackage{multirow}
\usepackage{hyperref}
\usepackage{xcolor}
\usepackage{orcidlink}

\title{thaulab@EEUCA 2026: Who Said What to Whom? A Targeting-Aware Neural-Symbolic Pipeline for Gaming Toxicity Detection}

\author{Anmol Guragain\orcidlink{0009-0009-8491-8663}$^*$, 
        Marcos Estecha Garitagoitia, 
        Luis Fernando D'Haro Enríquez,
        Ricardo Córdoba \\
  ETSI de Telecomunicación, Universidad Politécnica de Madrid, Madrid, Spain \\
  $^*$\texttt{anmol.g@upm.es} \ (Corresponding author)}

\begin{document}
\maketitle
\begin{abstract}
This paper describes our system for the EEUCA 2026 Shared Task on toxicity classification in gaming chat.
We implement a three-stage pipeline combining an ensemble of two compact transformers (DeBERTa-v3-base, 184M; XLM-RoBERTa-base, 278M) with a Linguistically-Informed Mediator (LIM) that resolves inter-model disagreements through corpus-backed lexical normalization, class-conditional unigram scoring, multilingual profanity detection, and agentive targeting analysis grounded in speech act theory.
The LIM specifically targets the minority classes (Hate \& Harassment, Threats, and Extremism), which are the most safety-critical categories in real-world gaming moderation.
To address the extreme class imbalance (1{,}450:1 Non-toxic to Extremism ratio), we introduce a two-stage data augmentation strategy using only the provided training data.
Our system achieves a Macro F1 of 0.6441 and accuracy of 0.9062 on the official test set, ranking 3rd in Macro F1 and 1st in accuracy among all teams.
The proposed pipeline is domain-portable: adapting to other gaming platforms requires substituting only the game-specific entity lexicon.
Code is publicly available at \url{https://github.com/Anmol2059/thaulab\_EEUCA}.
\end{abstract}

\section{Introduction}
\label{sec:intro}

Online gaming platforms host millions of real-time text interactions daily, and toxic behavior in these environments has been linked to serious consequences including cyberbullying, psychological harm, and player attrition \cite{parihar2021hate}.
A recent systematic review of 64 studies confirms that cyberbullying in multiplayer games is associated with anxiety, depression, and social withdrawal \cite{hu2025player}, and empirical evidence shows that toxic behavior propagates virally among teammates (exposure to toxic teammates increases a player's own toxicity likelihood by up to 30$\times$), amplifying its reach when left undetected \cite{morrier2024uncovering}.

The EEUCA 2026 Shared Task on Gaming Toxicity \cite{thapa2026toxicity,hurriyetouglu2026eeuca} introduces a six-class classification benchmark derived from World of Tanks chat logs \cite{naseem2025gametox}, annotated following the directed/undirected hate speech framework of \citet{bhandari2023crisishatemm}.
The dataset poses three key challenges: extreme class imbalance (81.0\% Non-toxic vs.\ 0.06\% Extremism), multilingual content spanning 10+ languages, and domain-specific lexical ambiguity where violent vocabulary (``kill'', ``destroy'') carries non-violent illocutionary force.

We implement a three-stage system combining neural ensemble classification with a rule-based Linguistically-Informed Mediator (LIM), following evidence that logical rules provide complementary signal to neural hate speech classifiers \cite{clarke2023rule,awasthilearning}.
Our contributions are:
(1)~a two-stage augmentation strategy (confusion-pair-driven and contrastive boundary generation) that improves Macro F1 by +9.7\% relative using only the provided data;
(2)~a LIM module grounded in speech act theory \cite{austin1962things,searle1969speech} that resolves ensemble disagreements through four interpretable, corpus-backed components;
(3)~empirical evidence that even multilingual transformers exhibit residual blind spots on domain-specific non-Latin profanity (22.6\% of Hate \& Harassment contains Cyrillic);
and (4)~demonstration that general-purpose toxicity models (toxic-bert) fail catastrophically in the gaming domain (Macro F1 = 0.3154), showing that gaming chat is a distinct linguistic register that requires domain-specific handling.

\paragraph{Related work.}
Recent NLP approaches to gaming toxicity include domain-adaptive pretraining of RoBERTa with match metadata for DOTA~2 and Call of Duty \cite{schurgerfoy2025contextaware}, and hybrid architectures combining LLM-generated embeddings with lightweight classifiers for Twitch moderation \cite{ansari2026toxitwitch}.
For class imbalance in hate speech detection, \citet{zhang2024study} show that focal loss \cite{lin2017focal} consistently yields peak performance, motivating our loss function choice.
LLM-based data augmentation has proven effective for hate speech minority classes \cite{li2026toxigan}, supporting our two-stage augmentation strategy (\S\ref{sec:augmentation}).
The GameTox dataset \cite{naseem2025gametox} additionally provides intent and slot filling annotations, but these labels were not released for the shared task, limiting participants to the six-class toxicity schema.
Annotation disagreement is a recognized challenge in hate speech classification \cite{dehghan2025dealing,bhandari2023crisishatemm}; we quantify its extent in this dataset in \S\ref{sec:noise}.

\section{Task, Dataset, and Augmentation}
\label{sec:task}

The shared task \cite{thapa2026toxicity,hurriyetouglu2026eeuca} requires classifying World of Tanks chat into six categories: \textbf{Non-toxic}~(0), \textbf{Insults}~(1), \textbf{Other Offensive}~(2), \textbf{Hate \& Harassment}~(3), \textbf{Threats}~(4), and \textbf{Extremism}~(5), evaluated by Macro F1.
Table~\ref{tab:augmentation} shows the class distribution; the dataset contains 42{,}959 training, 5{,}367 validation, and 5{,}375 test samples with extreme imbalance (Non-toxic to Extremism ratio of 1{,}450:1).
Messages are very short (median 2--4 tokens) and 6.9\% contain Cyrillic script, rising to 22.6\% in Hate \& Harassment.

\subsection{Two-Stage Augmentation}
\label{sec:augmentation}

We augment minority classes without external data using a two-stage strategy (Figure~\ref{fig:augmentation}; prompt templates in Appendix~\ref{app:augmentation_prompt}).

\textbf{Stage A} uses a seed model (M0, DeBERTa-v3-base trained on original data) to identify confused class pairs: for each validation sample, we record the two highest-probability classes from M0 and flag the pair as a confusion boundary when the second-highest probability exceeds 0.15 (set empirically), indicating non-trivial model uncertainty between the two classes \cite{swayamdipta2020dataset}.
This threshold was selected based on validation-set Macro F1 evaluated across candidate values $\{0.10, 0.15, 0.20, 0.25\}$; 0.15 maximized minority-class recall without introducing excess noise into the augmentation pool, as lower values produced near-duplicate confusion pairs while higher values missed meaningful boundary cases.
Claude Opus 4.6 then generates synthetic samples targeting these confusion boundaries.
This yields augmentation for Other Offensive, Hate \& Harassment, and Threats.

\textbf{Stage B} addresses Extremism ($n{=}24$) and supplemental Threats ($n{=}60$), which are too rare for confusion-pair analysis.
We apply contrastive boundary augmentation: (1)~mine class-discriminative tokens at $\geq 5\times$ frequency ratio using $P(c \mid w)$, the fraction of training messages containing word $w$ that belong to class $c$ (the same statistic used in the LIM's unigram scoring, \S\ref{sec:keyword}); (2)~generate cross-lingual variants and unmask leet-speak (e.g., \texttt{naz1}$\rightarrow$\texttt{nazi}); (3)~verify that generated samples fall within valid similarity bounds to real training data via cosine similarity in a TF-IDF subspace restricted to the mined discriminative vocabulary.

Table~\ref{tab:augmentation} shows the result: 631 synthetic samples, improving Macro F1 by +9.7\% relative over M0.

\begin{figure}[t]
\centering
\includegraphics[width=\columnwidth]{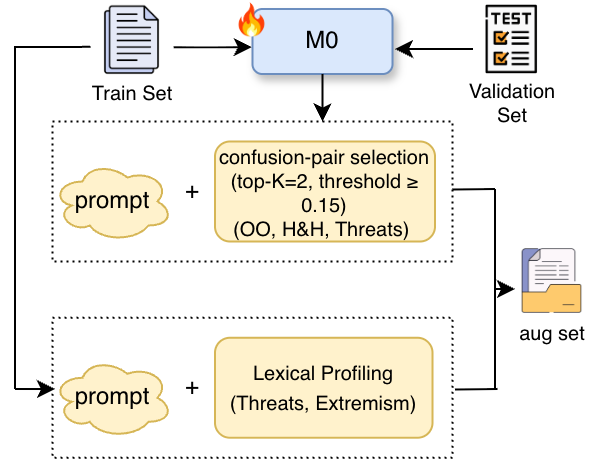}
\caption{Two-stage augmentation. Stage A: confusion-pair-driven. Stage B: contrastive boundary with lexical profiling.}
\label{fig:augmentation}
\end{figure}

\begin{table}[t]
\centering
\small
\begin{tabular}{lrrr}
\toprule
\textbf{Class} & \textbf{Original} & \textbf{Added} & \textbf{Final} \\
\midrule
Non-toxic          & 34{,}797 &   0 & 34{,}797 \\
Insults            & 5{,}925  &   0 & 5{,}925 \\
Other Offensive    & 1{,}874  &  34 & 1{,}908 \\
Hate \& Harassment & 279      & 155 & 434 \\
Threats            & 60       & 235 & 295 \\
Extremism          & 24       & 207 & 231 \\
\midrule
\textbf{Total}     & 42{,}959 & 631 & 43{,}590 \\
\bottomrule
\end{tabular}
\caption{Training set class distribution before and after augmentation.}
\label{tab:augmentation}
\end{table}

\section{System Architecture}
\label{sec:system}

Our system is a three-stage pipeline (Figure~\ref{fig:pipeline}); hyperparameters for all models are in Appendix~\ref{app:hyperparams}.

\begin{figure}[t]
\centering
\includegraphics[width=\columnwidth]{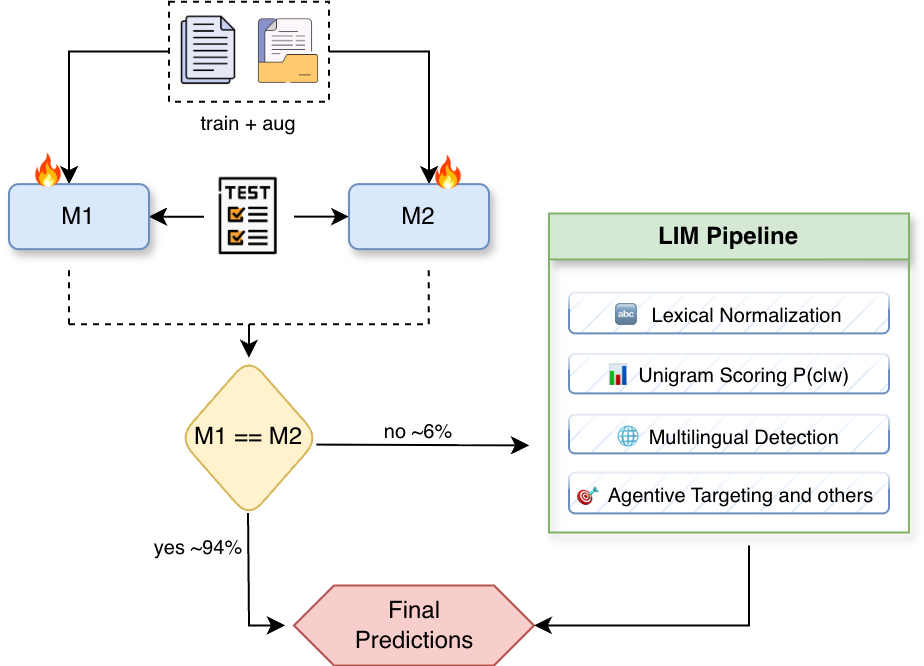}
\caption{System pipeline. M1 (DeBERTa) and M2 (XLM-R) produce predictions; agreements ($\sim$94\%) are accepted, disagreements ($\sim$6\%) are refined by the LIM.}
\label{fig:pipeline}
\end{figure}

\subsection{Stage 1: Model Exploration and Selection}
\label{sec:base}

We explored four base-sized transformers, all trained with focal loss \cite{lin2017focal} ($\gamma{=}2.0$, $\alpha{=}\text{None}$).
Table~\ref{tab:models} summarizes results.
We select M1 (DeBERTa-v3-base \cite{he2023debertav3}, 184M, highest F1 after augmented training) and M2 (XLM-RoBERTa \cite{conneau2020xlmroberta}, 278M, complementary multilingual coverage) for the ensemble.
M3 (BERT-base \cite{devlin2019bert}, 110M) served as a development baseline.
M4 (toxic-bert \cite{hanu2020detoxify}, 110M, frozen backbone + MLP) achieved only 0.3154 Macro F1 despite toxicity-specific pre-training, demonstrating that general-domain toxicity representations do not transfer to gaming contexts, where violent vocabulary is routinely non-toxic and multilingual slang is pervasive.
This is part of what the LIM's domain-specific linguistic rules are designed to address.

\begin{table}[t]
\centering
\small
\begin{tabular}{lcccc}
\toprule
\textbf{Model} & \textbf{F1} & \textbf{Acc} & \textbf{Prec} & \textbf{Rec} \\
\midrule
M0: DeBERTa (seed)  & .5234 & .8945 & .5463 & .5156 \\
M1: DeBERTa (aug.)   & .5742 & .8979 & .6042 & .5538 \\
M2: XLM-RoBERTa      & .5613 & .8910 & .5670 & .5730 \\
M3: BERT-base         & .5439 & .8947 & .5252 & .5811 \\
M4: toxic-bert        & .3154 & .7518 & .3187 & .5497 \\
\bottomrule
\end{tabular}
\caption{Individual model results. M0/M1 share the DeBERTa architecture (seed vs.\ augmented). M1 and M2 form the final ensemble.}
\label{tab:models}
\end{table}

\subsection{Stage 2: Agreement-Based Fusion}
\label{sec:fusion}

When M1 and M2 agree ($\sim$94\%), we accept the consensus.
For the $\sim$6\% disagreements, we adopt the prediction with higher softmax probability as the initial estimate and route to the LIM.

\subsection{Stage 3: Linguistically-Informed Mediator}
\label{sec:lim}

The LIM refines disagreement predictions through four sequential components.
It combines neural and symbolic processing: the ensemble captures distributional semantics, while the LIM encodes domain-specific linguistic facts that neural models cannot reliably learn from limited minority-class data.
Every LIM decision traces back to a specific rule and corpus statistic, making it auditable.

\subsubsection{Corpus-Backed Lexical Normalization}
\label{sec:exact}

We normalize test messages (lowercase, strip punctuation, collapse expressive lengthening \cite{brody2011coolool}: ``hahaha''$\rightarrow$``haha'') and perform exact-match lookup against train+val.
Matches with $\geq 2$ occurrences and $\geq 60\%$ majority agreement adopt the majority label.
These conservative thresholds directly reflect the annotation noise quantified in \S\ref{sec:noise}.

\subsubsection{Class-Conditional Unigram Scoring}
\label{sec:keyword}

Inspired by the token-level analysis of \citet{naseem2025gametox}, we compute $P(c \mid w) = \frac{n(w,c)}{n(w)}$ for each word $w$ in the training vocabulary, where $c$ is a class label, $n(w,c)$ is the number of messages containing $w$ in class $c$, and $n(w)$ is the total count, effectively a unigram Na\"{i}ve Bayes estimate.
Words exceeding a precision threshold $P(c \mid w) \geq 0.80$ with sufficient support ($n(w) \geq 5$) serve as high-confidence minority-class indicators \cite{wiegand2018inducing}.
For instance, identity-based slurs consistently map to H\&H ($P{=}1.00$), while ``kys'' maps to Threats and leet-speak variants like ``naz1'' to Extremism.
Overrides apply only toward safety-critical classes (3--5: H\&H, Threats, Extremism), prioritizing precision to avoid false escalation.

\subsubsection{Multilingual Profanity Detection}
\label{sec:multilingual}

While M2 (XLM-RoBERTa) handles multilingual tokenization, our validation analysis revealed that \emph{both} M1 and M2 still misclassify domain-specific non-Latin profanity, particularly terms rare even in XLM-R's 100-language pre-training.
We applied the same statistics to non-Latin tokens, flagging words where $P(\text{toxic} \mid w) = 1 - P(\text{Non-toxic} \mid w) \geq 0.80$ but both models predicted Non-toxic.
This yielded an empirically-validated multilingual lexicon organized by language family: East Slavic (Russian, Ukrainian), West Slavic (Polish, Czech), and other (Turkish, Hungarian, German).
Reclassification follows targeting: player-directed $\rightarrow$ Insults; game-directed $\rightarrow$ Other Offensive.

\subsubsection{Agentive Targeting and Pragmatic Refinement}
\label{sec:targeting}

Drawing on speech act theory \cite{austin1962things,searle1969speech}, we formalize the targeting function $\tau(m)$.
Let $T$ denote the set of tokens flagged as toxic by the preceding LIM components (unigram scoring and multilingual detection).
For a message $m$ containing a toxic token $t \in T$:
\begin{equation}
\tau(m) = \begin{cases}
\textsc{Other\text{-}dir} & \text{if } \exists\, p \in P_2 : p \prec t \\
\textsc{Self\text{-}dir} & \text{if } \exists\, p \in P_1 : p \prec t \\
\textsc{Entity\text{-}dir} & \text{if } \exists\, e \in E : e \prec t \\
\textsc{Untargeted} & \text{otherwise}
\end{cases}
\label{eq:targeting}
\end{equation}
where $P_2$/$P_1$ are second/first-person pronoun sets (English and Russian), $E$ is a Game-Specific Entity (GSE) lexicon covering vehicles (400+ tanks), mechanics (\emph{rng}, \emph{arty}, \emph{cap}), map locations (\emph{Himmelsdorf}, \emph{hill}, \emph{banana}), and game roles (\emph{light}, \emph{heavy}, \emph{TD}), and $\prec$ denotes linear precedence in the message.
Table~\ref{tab:targeting} maps targeting types to labels.

\begin{table}[h!]
\centering
\small
\begin{tabular}{lll}
\toprule
\textbf{$\tau(m)$} & \textbf{Signal} & \textbf{$\hat{y}$} \\
\midrule
\textsc{Other-dir}  & $P_2 + T$ (\emph{you} + insult)  & Insults \\
\textsc{Self-dir}   & $P_1 + T$ (\emph{I} + insult)    & Non-toxic \\
\textsc{Entity-dir} & $E + T$ (\emph{arty} + profanity) & Other Off. \\
\textsc{Untargeted} & $T$ only                           & Non-toxic \\
\bottomrule
\end{tabular}
\caption{Targeting function $\tau(m)$. $E$ = GSE lexicon. Based on speech act theory \cite{searle1969speech}.}
\label{tab:targeting}
\end{table}

This component also applies censored-text recovery (\textsc{[GSE]} + \texttt{[***]} $\rightarrow$ Non-toxic) and implicit word sense disambiguation: ``kill that Tiger'' (GSE $\rightarrow$ Non-toxic) vs.\ ``kill yourself'' (person $\rightarrow$ Threats) \cite{firth1957synopsis}.

\section{Results and Discussion}
\label{sec:results}

\begin{table}[t]
\centering
\small
\begin{tabular}{lcccc}
\toprule
\textbf{System} & \textbf{F1} & \textbf{Acc} & \textbf{Prec} & \textbf{Rec} \\
\midrule
Best single (M1)         & .5742 & .8979 & .6042 & .5538 \\
M1+M2 ensemble           & .6032 & .9059 & .5713 & .6579 \\
\midrule
\quad+ Lex.\ norm.       & .6107 & .9057 & .5782 & .6591 \\
\quad+ Unigram scoring   & .6221 & .9044 & .5964 & .6559 \\
\quad+ Multilingual      & .6256 & .9047 & .5970 & .6626 \\
\quad+ Targeting \& ref. & \textbf{.6441} & \textbf{.9062} & \textbf{.6334} & \textbf{.6601} \\
\bottomrule
\end{tabular}
\caption{Incremental ablation. Top: models. Bottom: LIM components applied to $\sim$6\% disagreements. Full LIM includes targeting, censored-text recovery, and boundary enforcement.}
\label{tab:main}
\end{table}

Table~\ref{tab:main} shows incremental results on the official test set. The ensemble improves over the best single model through complementary coverage, and the LIM further refines the $\sim$6\% disagreements, with the largest contribution from unigram scoring (Table~\ref{tab:main}).
The LIM's impact is concentrated in safety-critical minority classes, where high-precision corrections ensure that identity-based hate, threats, and extremist content are not missed by the neural ensemble.

Our system ranks 3rd in Macro F1 (0.6441) but achieves the \textbf{highest accuracy} (0.9062) among all participating teams, indicating the fewest total errors; the F1 gap to the top-ranked systems (0.7041, 0.6725) is concentrated in minority class recall.
\textbf{Annotation noise} accounts for part of this ceiling: 340 unique messages carry conflicting labels across 7{,}416 training samples (17.3\%), with the Non-toxic $\leftrightarrow$ Insults boundary alone responsible for 6{,}455 conflicting samples, reflecting the same ambiguity between a playful insult and a genuine attack that makes this boundary the hardest to learn.
\textbf{Multilingual blind spots} persist even with XLM-RoBERTa: domain-specific Cyrillic profanity is concentrated at 22.6\% of H\&H messages (3.5$\times$ the dataset average), reflecting the prevalence of Russian and Ukrainian identity-based slurs that fall outside standard multilingual pre-training corpora.
\textbf{Domain gap}: the toxic-bert result (F1 = 0.3154 vs.\ 0.5439 for vanilla BERT) shows that Twitter/Reddit toxicity pre-training actively hurts gaming performance by associating GSE terms with toxicity.

\label{sec:noise}

\label{sec:conclusion}

We presented \textbf{thaulab}'s system for the EEUCA 2026 GameTox Shared Task, achieving Macro F1 of 0.6441 (3rd) and the highest accuracy (0.9062) using exclusively base-sized models and no external data.
The pipeline is adaptive by design: augmentation targets the boundaries the model struggles with, and the LIM concentrates corrections on Extremism, Threats, and H\&H, categories where misclassification causes real harm beyond any leaderboard metric.
The three-stage framework generalizes to other gaming platforms with only GSE lexicon substitution.
Key findings: (1)~symbolic mediation on ensemble disagreements improves safety-critical minority-class detection; (2)~multilingual transformers retain blind spots on domain-specific profanity; (3)~agentive targeting distinguishes toxic intent from benign game communication; (4)~general toxicity models fail in gaming contexts.

\newpage
\section*{Limitations}

The LIM relies on simple symbolic methods (lexicon lookup, unigram statistics, pronoun-based targeting) rather than learned components. While this trades off flexibility for explainability, since every correction traces back to a specific corpus statistic or linguistic pattern, it likely leaves performance on the table. We set strict thresholds throughout (e.g., $P(c \mid w) \geq 0.80$, majority agreement $\geq 60\%$) and have not yet explored relaxing them per class or per confidence region. The four LIM components are applied in a fixed sequential order without searching over alternative orderings, and the LIM currently operates only on the $\sim$6\% of samples where M1 and M2 disagree; extending symbolic mediation to the full prediction set could improve coverage but risks false positives at the noisy majority-class boundaries. On the augmentation side, all synthetic data was generated using Claude Opus 4.6, a safety-aligned model that may undergenerate realistic toxic content due to its content filtering; less restricted models (e.g., Grok) could potentially yield more naturalistic minority-class samples, particularly for Extremism where only 24 original samples exist. Finally, the multilingual lexicon is manually curated and incomplete for unrepresented languages, the GSE lexicon is specific to World of Tanks, targeting analysis cannot resolve implicit targeting or zero pronouns in pro-drop languages, and we could not experiment with larger backbone models due to time and compute constraints.

\section*{Ethics Statement}

This work classifies toxic language including slurs and profanity in multiple languages.
Examples are reported solely for scientific reproducibility.
Our multilingual profanity lexicon is compiled for research purposes only.


\section*{Acknowledgements}

This work is supported by the project BRAINS (PID2024-155948OB-C52), funded by MCIN/AEI/10.13039/501100011033 and, as appropriate, by ``ERDF A way of making Europe'' and by the ``European Union''; and by the Program \emph{Actividades I+D Procesos Humanos y Sociales} of the Comunidad de Madrid, Spain, under Project PHS2024/PH-HUM-52 (Innovatrad-CM).

\bibliography{custom}

@inproceedings{thapa2026toxicity,
  title={Understanding Toxic Behavior in Gaming Communities Using {AI} to Promote Healthier Digital Spaces},
  author={Thapa, Surendrabikram and Shiwakoti, Shuvam and Shah, Siddhant Bikram and Rauniyar, Kritesh and Thapa, Laxmi and Adhikari, Surabhi and Johnson, Kristina T. and H{\"u}rriyeto{\u{g}}lu, Ali and Tanev, Hristo and Naseem, Usman},
  booktitle={Proceedings of the 9th Workshop on Event Extraction and Understanding: Challenges and Applications (EEUCA)},
  year={2026}
}

@inproceedings{hurriyetouglu2026eeuca,
  title={Overview of the Workshop on Event Extraction and Understanding: Challenges and Applications},
  author={H{\"u}rriyeto{\u{g}}lu, Ali and Thapa, Surendrabikram and Tanev, Hristo and Thapa, Laxmi and Adhikari, Surabhi},
  booktitle={Proceedings of the 9th Workshop on Event Extraction and Understanding: Challenges and Applications (EEUCA)},
  year={2026}
}

@inproceedings{naseem2025gametox,
  title={Gametox: A Comprehensive Dataset and Analysis for Enhanced Toxicity Detection in Online Gaming Communities},
  author={Naseem, Usman and Shiwakoti, Shuvam and Shah, Siddhant Bikram and Thapa, Surendrabikram and Zhang, Qi},
  booktitle={Proceedings of the 2025 Conference of the Nations of the Americas Chapter of the Association for Computational Linguistics: Human Language Technologies (Volume 2: Short Papers)},
  pages={440--447},
  year={2025}
}

@inproceedings{bhandari2023crisishatemm,
  title={{CrisisHateMM}: Multimodal Analysis of Directed and Undirected Hate Speech in Text-Embedded Images from {Russia-Ukraine} Conflict},
  author={Bhandari, Aashish and Shah, Siddhant B and Thapa, Surendrabikram and Naseem, Usman and Nasim, Mehwish},
  booktitle={Proceedings of the IEEE/CVF Conference on Computer Vision and Pattern Recognition},
  pages={1994--2003},
  year={2023}
}

@inproceedings{parihar2021hate,
  title={Hate Speech Detection Using Natural Language Processing: Applications and Challenges},
  author={Parihar, Anil Singh and Thapa, Surendrabikram and Mishra, Sushruti},
  booktitle={2021 5th International Conference on Trends in Electronics and Informatics (ICOEI)},
  pages={1302--1308},
  year={2021},
  organization={IEEE}
}

@article{he2023debertav3,
  title={{DeBERTaV3}: Improving {DeBERTa} using {ELECTRA}-Style Pre-Training with Gradient-Disentangled Embedding Sharing},
  author={He, Pengcheng and Gao, Jianfeng and Chen, Weizhu},
  journal={arXiv preprint arXiv:2111.09543},
  year={2023}
}

@inproceedings{devlin2019bert,
  title={{BERT}: Pre-training of Deep Bidirectional Transformers for Language Understanding},
  author={Devlin, Jacob and Chang, Ming-Wei and Lee, Kenton and Toutanova, Kristina},
  booktitle={Proceedings of the 2019 Conference of the North American Chapter of the Association for Computational Linguistics: Human Language Technologies (Volume 1: Long and Short Papers)},
  pages={4171--4186},
  year={2019}
}

@inproceedings{lin2017focal,
  title={Focal Loss for Dense Object Detection},
  author={Lin, Tsung-Yi and Goyal, Priya and Girshick, Ross and He, Kaiming and Doll{\'a}r, Piotr},
  booktitle={Proceedings of the IEEE International Conference on Computer Vision (ICCV)},
  pages={2980--2988},
  year={2017}
}

@article{conneau2020xlmroberta,
  title={Unsupervised Cross-lingual Representation Learning at Scale},
  author={Conneau, Alexis and Khandelwal, Kartikay and Goyal, Naman and Chaudhary, Vishrav and Wenzek, Guillaume and Guzm{\'a}n, Francisco and Grave, Edouard and Ott, Myle and Zettlemoyer, Luke and Stoyanov, Veselin},
  journal={Proceedings of the 58th Annual Meeting of the Association for Computational Linguistics},
  pages={8440--8451},
  year={2020}
}

@article{hu2025player,
  title={Player versus Player: A Systematic Review of Cyberbullying in Multiplayer Online Games},
  author={Hu, Yunhao and Evelyn, Sophie and Clancy, Elizabeth M.},
  journal={Computers in Human Behavior},
  year={2025},
  publisher={Elsevier}
}

@misc{morrier2024uncovering,
  title={Uncovering the Viral Nature of Toxicity in Competitive Online Video Games},
  author={Morrier, Jacob and Mahmassani, Amine and Alvarez, R. Michael},
  year={2024},
  eprint={2410.00978},
  archivePrefix={arXiv},
  primaryClass={cs.CY}
}

@misc{schurgerfoy2025contextaware,
  title={Context-Aware Toxicity Detection in Multiplayer Games: Integrating Domain-Adaptive Pretraining and Match Metadata},
  author={Schurger-Foy, Adrien and Kocielnik, Rafal Dariusz and Gulcehre, Caglar and Alvarez, R. Michael},
  year={2025},
  eprint={2504.01534},
  archivePrefix={arXiv},
  primaryClass={cs.CL}
}

@misc{ansari2026toxitwitch,
  title={{ToxiTwitch}: Toward Emote-Aware Hybrid Moderation for Live Streaming Platforms},
  author={Ansari, Baktash and Martin, Elias and Mashhadi, Afra},
  year={2026},
  eprint={2601.15605},
  archivePrefix={arXiv},
  primaryClass={cs.HC}
}

@inproceedings{zhang2024study,
  title={A Study of the Class Imbalance Problem in Abusive Language Detection},
  author={Zhang, Yaqi and Hangya, Viktor and Fraser, Alexander},
  booktitle={Proceedings of the 8th Workshop on Online Abuse and Harms (WOAH 2024)},
  year={2024},
  publisher={Association for Computational Linguistics}
}

@inproceedings{li2026toxigan,
  title={{ToxiGAN}: Toxic Data Augmentation via {LLM}-Guided Directional Adversarial Generation},
  author={Li, Peiran and Fillies, Jan and Paschke, Adrian},
  booktitle={Proceedings of the 18th Conference of the European Chapter of the Association for Computational Linguistics (EACL)},
  year={2026}
}

@misc{dehghan2025dealing,
  title={Dealing with Annotator Disagreement in Hate Speech Classification},
  author={Dehghan, Somaiyeh and Sen, Mehmet Umut and Yanikoglu, Berrin},
  year={2025},
  eprint={2502.08266},
  archivePrefix={arXiv},
  primaryClass={cs.CL}
}

@book{austin1962things,
  title={How to Do Things with Words},
  author={Austin, John L.},
  year={1962},
  publisher={Oxford University Press}
}

@book{searle1969speech,
  title={Speech Acts: An Essay in the Philosophy of Language},
  author={Searle, John R.},
  year={1969},
  publisher={Cambridge University Press}
}

@article{firth1957synopsis,
  title={A Synopsis of Linguistic Theory, 1930--1955},
  author={Firth, John R.},
  journal={Studies in Linguistic Analysis},
  pages={1--32},
  year={1957},
  publisher={Blackwell}
}

@inproceedings{brody2011coolool,
  title={Cooooooooooooooollllllllllllll!!!!!!!!!!!!!!: Using Word Lengthening to Detect Sentiment in Microblogs},
  author={Brody, Samuel and Diakopoulos, Nicholas},
  booktitle={Proceedings of the 2011 Conference on Empirical Methods in Natural Language Processing},
  pages={562--570},
  year={2011}
}

@inproceedings{clarke2023rule,
  title={Rule By Example: Harnessing Logical Rules for Explainable Hate Speech Detection},
  author={Clarke, Christopher and Hall, Matthew and Mittal, Gaurav and Yu, Ye and Sajeev, Sandra and Mars, Jason and Chen, Mei},
  booktitle={Proceedings of the 61st Annual Meeting of the Association for Computational Linguistics (Volume 1: Long Papers)},
  pages={364--376},
  year={2023}
}

@inproceedings{awasthilearning,
  title={Learning from Rules Generalizing Labeled Exemplars},
  author={Awasthi, Abhijeet and Ghosh, Sabyasachi and Goyal, Rasna and Sarawagi, Sunita},
  booktitle={International Conference on Learning Representations},
  year={2020}
}

@inproceedings{swayamdipta2020dataset,
  title={Dataset cartography: Mapping and diagnosing datasets with training dynamics},
  author={Swayamdipta, Swabha and Schwartz, Roy and Lourie, Nicholas and Wang, Yizhong and Hajishirzi, Hannaneh and Smith, Noah A and Choi, Yejin},
  booktitle={Proceedings of the 2020 Conference on Empirical Methods in Natural Language Processing (EMNLP)},
  pages={9275--9293},
  year={2020}
}

@inproceedings{wiegand2018inducing,
  title={Inducing a lexicon of abusive words--a feature-based approach},
  author={Wiegand, Michael and Ruppenhofer, Josef and Schmidt, Anna and Greenberg, Clayton},
  booktitle={Proceedings of the 2018 Conference of the North American Chapter of the Association for Computational Linguistics: Human Language Technologies, Volume 1 (Long Papers)},
  pages={1046--1056},
  year={2018}
}

@misc{hanu2020detoxify,
  title={Detoxify},
  author={Hanu, Laura and {Unitary team}},
  howpublished={https://huggingface.co/unitary/toxic-bert},
  year={2020}
}

\newpage
\newpage
\appendix

\section{Hyperparameter Configuration}
\label{app:hyperparams}

Table~\ref{tab:hyperparams} lists the model-specific training configuration for all four architectures.
The pre-trained checkpoints are \texttt{microsoft/deberta-v3-base} (M0/M1), \texttt{xlm-roberta-base} (M2), \texttt{bert-base-uncased} (M3), and \texttt{unitary/toxic-bert} (M4).
All models share the same base settings: focal loss with focusing parameter $\gamma = 2.0$ and no class balancing ($\alpha = \text{None}$), maximum sequence length of 64 tokens, batch size 32, 5 training epochs, AdamW optimizer with weight decay 0.01, and random seed 42.

While M2 and M3 natively instantiate in single-precision (\texttt{float32}), the DeBERTa-v3 checkpoints (M0/M1) are natively stored in half-precision (\texttt{float16}).
We observed that fine-tuning DeBERTa-v3 in \texttt{float16} resulted in catastrophic gradient collapse (\texttt{NaN} loss) during the initial training steps, a known instability caused by arithmetic overflow within DeBERTa's Disentangled Attention matrices, where intermediate activation values exceed the \texttt{float16} maximum representable limit.
To resolve this, we explicitly upcast the DeBERTa weights to \texttt{float32} during initialization, providing sufficient numerical stability for the attention mechanism to converge.
We additionally tested $\gamma = 2.5$, dynamic $\alpha$ (inverse class frequency), and a two-stage hierarchical approach (binary toxic/non-toxic classification followed by fine-grained 6-class prediction within the toxic branch).
All alternatives yielded marginal differences ($\Delta$ Macro F1 $< 0.005$), so we standardized the simplest configuration for reproducibility across all architectures.

\begin{table*}[ht]
\centering
\small
\begin{tabular}{lcccc}
\toprule
\textbf{Parameter} & \textbf{M0/M1 (DeBERTa-v3)} & \textbf{M2 (XLM-RoBERTa)} & \textbf{M3 (BERT)} & \textbf{M4 (toxic-bert)} \\
\midrule
Total Parameters       & 184M            & 278M            & 110M            & 110M \\
Trainable Parameters   & 184M (full)     & 278M (full)     & 110M (full)     & $\sim$200K (MLP only) \\
Training Paradigm      & Full fine-tune  & Full fine-tune  & Full fine-tune  & Frozen backbone + MLP \\
Learning Rate          & $1 \times 10^{-5}$ & $2 \times 10^{-5}$ & $2 \times 10^{-5}$ & $1 \times 10^{-3}$ (MLP) \\
Warmup Steps           & 10\% of total   & 0               & 0               & 0 \\
Weight Initialization  & \texttt{float32} (upcast) & \texttt{float32} (native) & \texttt{float32} (native) & \texttt{float32} (native) \\
\bottomrule
\end{tabular}
\caption{Model-specific training hyperparameters. M0 and M1 share the identical configuration; M0 is trained on the original training set, M1 on the augmented set.}
\label{tab:hyperparams}
\end{table*}

\section{LIM Component Details}
\label{app:thresholds}

Table~\ref{tab:thresholds} lists all LIM thresholds, selected on the validation set and held fixed during test evaluation.
The lexical normalization majority threshold was set at 60\% rather than 50\% because lower values introduced false corrections at the noisy Insults $\leftrightarrow$ Other Offensive boundary.
The unigram precision cutoff of $P \geq 0.80$ was chosen because at $P \geq 0.70$, ambiguous terms (e.g., ``monkey'' at $P(\text{H\&H}){=}0.75$) triggered false positives; raising to 0.80 retains only unambiguous high-precision tokens.
These thresholds are intentionally strict for the competition setting and can be relaxed for higher-recall deployment.

\begin{table}[ht]
\centering
\small
\begin{tabular}{lll}
\toprule
\textbf{Component} & \textbf{Parameter} & \textbf{Value} \\
\midrule
Lex.\ Norm. & Min occurrences     & $\geq 2$ \\
             & Majority threshold  & $\geq 60\%$ \\
             & Normalization       & Lower, strip, \\
             &                     & dedup ($\geq 3$) \\
\midrule
Unigram      & $P(c \mid w)$ cutoff  & $\geq 0.80$ \\
             & Min support $n(w)$    & $\geq 5$ \\
             & Direction             & Minority $\uparrow$ only \\
             & Freq.\ ratio          & $\geq 5\times$ baseline \\
\midrule
Multilingual & Languages             & 10+ \\
             & Mining criterion      & $P(\text{toxic} \mid w) \geq 0.80$ \\
             &                       & + both models wrong \\
             & Reclassification      & Targeting-sensitive \\
\midrule
Targeting    & Pronoun sets          & EN, RU \\
             & GSE lexicon $E$       & Vehicles (400+), \\
             &                       & mechanics, maps, roles \\
             & Censored pattern      & [GSE] + [***] \\
\bottomrule
\end{tabular}
\caption{LIM thresholds. All selected on the validation set.}
\label{tab:thresholds}
\end{table}

Throughout the LIM, we enforce annotation-guideline boundaries \cite{naseem2025gametox}: identity-based slurs $\rightarrow$ H\&H; 2nd person + non-identity insult $\rightarrow$ Insults; profanity without personal targeting $\rightarrow$ Other Offensive; game callouts and GSE terms $\rightarrow$ Non-toxic; directed violence + personal target $\rightarrow$ Threats; political ideology and recruitment $\rightarrow$ Extremism.

\section{Augmentation Pipeline}
\label{app:augmentation_prompt}

\paragraph{Toxic vocabulary mining.}
Class-conditional unigram probabilities $P(c \mid w)$ are computed as described in \S\ref{sec:keyword}.
For augmentation, we additionally flag \emph{class-discriminative tokens} using the frequency ratio $\frac{n(w,c)/N_c}{n(w)/N} \geq 5$, where $N_c$ and $N$ are class and corpus sizes respectively; this yields a focused toxic vocabulary substantially smaller than the full $\sim$30K vocabulary, defining the TF-IDF subspace for similarity gating below.

Before computing any statistics, all text undergoes leet-speak normalization to unmask common obfuscation patterns prevalent in gaming chat.
The character substitution mappings are: \texttt{0}$\rightarrow$\texttt{o}, \texttt{1}$\rightarrow$\texttt{i}, \texttt{3}$\rightarrow$\texttt{e}, \texttt{4}$\rightarrow$\texttt{a}, \texttt{5}$\rightarrow$\texttt{s}, \texttt{7}$\rightarrow$\texttt{t}, \texttt{@}$\rightarrow$\texttt{a}, \texttt{\$}$\rightarrow$\texttt{s}.
This normalization is applied consistently in both the augmentation pipeline (for seed term selection and similarity verification) and the LIM (for unigram scoring and multilingual detection at inference time).

\paragraph{Cosine similarity gating in the toxic subspace.}
To verify that generated samples are linguistically consistent with real training data, we project both real and synthetic utterances into a \emph{toxic-only TF-IDF subspace}.
Rather than computing TF-IDF vectors over the full $\sim$30K vocabulary (which produces extremely sparse, high-dimensional vectors for short gaming messages of 2--4 tokens), we restrict the vocabulary to only the mined class-discriminative terms.
This projection substantially reduces dimensionality and eliminates the sparsity problem inherent in full-vocabulary TF-IDF for short texts.
Cosine similarity between each generated sample and its nearest real training neighbor in this subspace serves as a geometric filter: samples that fall below a minimum similarity threshold are rejected as out-of-distribution, while samples above a maximum threshold are rejected as near-duplicates of existing training data.
This dual-threshold approach ensures that generated samples are close enough to the training distribution to be realistic, yet sufficiently novel to provide genuine augmentation value.

\paragraph{Generation API configuration.}
All synthetic samples were generated using the Claude Opus 4.6 API (\texttt{claude-opus-4-6-20250514}).
We used a temperature of \texttt{1.0} to encourage lexical diversity across generated samples, \texttt{max\_tokens = 2048}, and \texttt{top\_p = 1.0} (no nucleus truncation).
No additional system-level parameters were set beyond the defaults; the full generation behavior is governed solely by the prompt templates below.
These settings are fixed across both Template A and Template B calls to ensure reproducibility.

\paragraph{Template A: Confusion-pair-driven generation.}
For classes identified through the seed model's (M0) prediction uncertainty, we provide the language model with the target class definition, representative seed examples from the training set, and the specific confused class pair that the model struggles with:

\begin{quote}
\small
\texttt{You are a data augmentation assistant for a toxicity classification dataset derived from World of Tanks in-game chat. Your task is to generate realistic synthetic chat messages for a specific toxicity class.}

\texttt{Target class: \{CLASS\_NAME\}}\\
\texttt{Class definition: \{CLASS\_DEFINITION\}}

\texttt{The class definitions follow the annotation guidelines from the GameTox dataset (Naseem et al., 2025):}\\
\texttt{- Hate and Harassment: Identity-based hate or harassment (racism, sexism, homophobia)}\\
\texttt{- Threats: Threats of violence, physical safety, terrorism, or doxxing}\\
\texttt{- Extremism: Extremist views, grooming/recruitment for extremist groups}\\
\texttt{- Insults and Flaming: Insults or attacks not based on identity}\\
\texttt{- Other Offensive: Offensive content not covered by the above categories}\\
\texttt{- Non-toxic: Neutral game communication}

\texttt{Seed examples from the training data:}\\
\texttt{\{SEED\_EXAMPLES\}}

\texttt{Confused with: \{CONFUSED\_CLASS\} (our classifier frequently confuses \{CLASS\_NAME\} with \{CONFUSED\_CLASS\})}

\texttt{Requirements:}\\
\texttt{1. Generate exactly 20 new chat messages that CLEARLY belong to \{CLASS\_NAME\} and NOT to \{CONFUSED\_CLASS\}.}\\
\texttt{2. Each message should be 1--8 words long (typical length in game chat).}\\
\texttt{3. Include common gaming abbreviations, slang, and informal spelling.}\\
\texttt{4. Include multilingual variants where appropriate (Russian, Polish, Turkish, German).}\\
\texttt{5. Each message must be unambiguously classifiable by a human annotator following the guidelines above.}\\
\texttt{6. Do NOT repeat or closely paraphrase the seed examples.}\\
\texttt{7. Output one message per line with no numbering or formatting.}
\end{quote}

\paragraph{Template B: Contrastive boundary augmentation.}
For extreme minority classes (Extremism with only $n = 24$ training samples, and supplemental Threats with $n = 60$) that are too rare to appear reliably in confusion-pair analysis, we provide discriminative keywords mined from the training set along with explicit instructions to generate boundary-proximal samples:

\begin{quote}
\small
\texttt{You are a data augmentation assistant for a toxicity classification dataset from World of Tanks in-game chat.}

\texttt{Target class: \{CLASS\_NAME\}}\\
\texttt{Class definition: \{CLASS\_DEFINITION\}}\\
\texttt{Adjacent (easily confused) class: \{ADJACENT\_CLASS\}}\\
\texttt{Adjacent class definition: \{ADJACENT\_DEFINITION\}}

\texttt{Discriminative keywords for \{CLASS\_NAME\} (statistically mined from training data, P(class|word) >= 0.80):}\\
\texttt{\{HIGH\_P\_KEYWORDS\}}

\texttt{Existing training examples of \{CLASS\_NAME\}:}\\
\texttt{\{SEED\_EXAMPLES\}}

\texttt{Requirements:}\\
\texttt{1. Generate exactly 20 new messages that belong to \{CLASS\_NAME\}.}\\
\texttt{2. CRITICAL: Messages must be CLOSE to the decision boundary with \{ADJACENT\_CLASS\}. They should be challenging to classify, but still clearly \{CLASS\_NAME\} according to the annotation guidelines.}\\
\texttt{3. Include cross-lingual variants (Russian, Polish, Turkish, German).}\\
\texttt{4. Vary message length (1--8 words).}\\
\texttt{5. Each message should be distinguishable from \{ADJACENT\_CLASS\} ONLY by the specific class-defining linguistic feature (e.g., identity targeting for H\&H vs. skill targeting for Insults, or political ideology for Extremism vs. identity hate for H\&H).}\\
\texttt{6. Do NOT repeat seed examples.}\\
\texttt{7. Output one message per line with no numbering.}
\end{quote}

\paragraph{Generation results and quality control.}
Template A yielded 34 Other Offensive, 155 Hate \& Harassment, and a portion of the Threats samples.
Template B yielded all 207 Extremism samples and supplemental Threats samples, for a combined total of 631 synthetic samples.
All generated samples underwent three quality control steps: (1) cosine similarity gating in the toxic TF-IDF subspace to reject out-of-distribution and near-duplicate generations; (2) exact and near-duplicate removal against the original training set to prevent data leakage; (3) manual spot-checking of a random 10\% subset for label consistency with the annotation guidelines.
Class definitions in both templates were drawn directly from the annotation guidelines of \citet{naseem2025gametox}.

\section{Annotation Noise Analysis}
\label{sec:noise}

A key challenge in the GameTox dataset is annotation inconsistency at class boundaries.
We identify 340 unique normalized messages that appear with conflicting labels across their multiple occurrences in the training set, collectively affecting 7{,}416 individual training samples (17.3\% of the dataset).
This inconsistency arises because identical text appears in different game sessions and receives different annotations each time.
For instance, a player typing ``wtf'' in one match may be reacting to an unfair death (Other Offensive), while in another match the same message is interpreted as a neutral exclamation (Non-toxic).
This is not a failure of multiple annotators disagreeing on a single instance; rather, it reflects the genuine context-dependence of short gaming messages.

Table~\ref{tab:noise_examples} shows representative examples.
The column $n$ indicates the total number of times that normalized message appears in the training set across all game sessions.
The label distribution shows the percentage of those $n$ occurrences assigned to each class.

\begin{figure}[ht]
\centering
\includegraphics[width=\columnwidth]{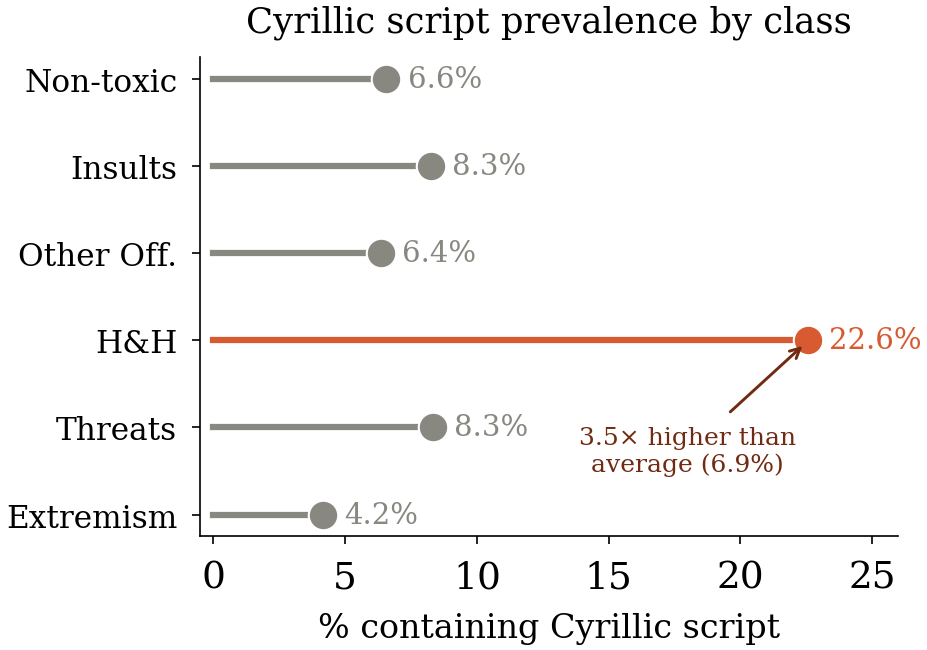}
\caption{Cyrillic-script prevalence by toxicity class. Hate \& Harassment contains 3.5$\times$ more Cyrillic content than the dataset average (6.9\%), indicating that non-Latin-script profanity is structurally concentrated in the most severe toxicity category.}
\label{fig:cyrillic}
\end{figure}

\begin{table}[ht]
\centering
\small
\begin{tabular}{lrl}
\toprule
\textbf{Message} & \textbf{$n$} & \textbf{Label Distribution} \\
\midrule
\texttt{gg}     & 2{,}703 & NT: 99.9\%, Ins: 0.1\% \\
\texttt{wtf}    & 208     & OO: 77.4\%, NT: 22.1\%, Ins: 0.5\% \\
\texttt{cap}    & 192     & NT: 96.4\%, OO: 2.1\%, Ins: 1.6\% \\
\texttt{arty}   & 189     & NT: 97.4\%, Ins: 2.6\% \\
\texttt{idiot}  & 101     & Ins: 94.1\%, NT: 5.0\%, OO: 1.0\% \\
\texttt{ffs}    & 55      & OO: 80.0\%, NT: 16.4\%, Ins: 3.6\% \\
\bottomrule
\end{tabular}
\caption{Annotation noise examples. $n$ = total occurrences in training data. The same normalized text receives different labels across game sessions. NT = Non-toxic, Ins = Insults, OO = Other Offensive.}
\label{tab:noise_examples}
\end{table}

\begin{figure}[ht]
\centering
\includegraphics[width=\columnwidth]{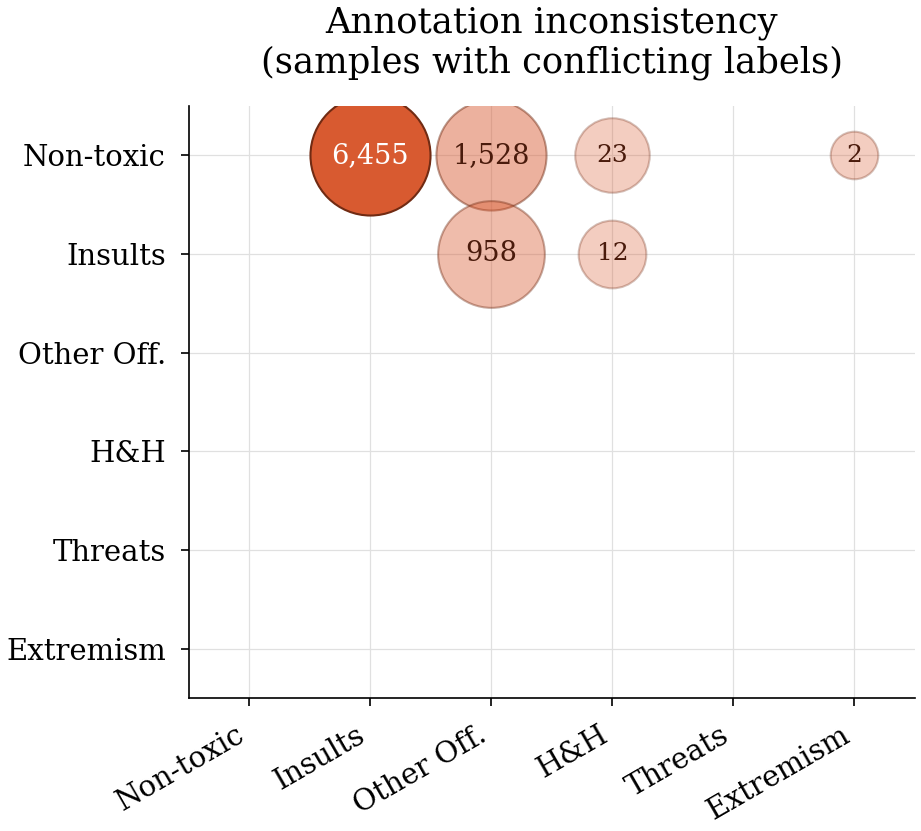}
\caption{Annotation inconsistency across all class pairs. Bubble size reflects the total number of training samples where identical normalized messages receive conflicting labels from the two classes. The Non-toxic $\leftrightarrow$ Insults boundary dominates at 6{,}455 conflicting samples, illustrating the context-dependent nature of short gaming messages.}
\label{fig:noise}
\end{figure}

Figure~\ref{fig:noise} visualizes the magnitude of annotation conflicts across all class pairs.
Each bubble represents a pair of classes; the bubble size and color intensity are proportional to the number of training samples where the same message receives labels from both classes.
The Non-toxic $\leftrightarrow$ Insults boundary dominates with 6{,}455 conflicting samples, reflecting the fundamental ambiguity between a playful insult and a genuine attack in gaming chat.
The Non-toxic $\leftrightarrow$ Other Offensive boundary (1{,}528 samples) and the Insults $\leftrightarrow$ Other Offensive boundary (958 samples) are the next most noisy.
Notably, minority class boundaries (involving H\&H, Threats, or Extremism) exhibit minimal noise, because the linguistic signals for these classes (identity-based slurs, directed violence, political ideology) are more distinctive and less context-dependent.

This noise pattern directly informs the LIM design: we use conservative thresholds ($\geq 60\%$ majority agreement) for the noisy majority-class boundaries, while applying more aggressive corrections for minority classes where annotation agreement is near-unanimous.

\section{Multilingual Content Analysis}
\label{app:multilingual}

Figure~\ref{fig:cyrillic} shows the proportion of Cyrillic-script messages by toxicity class.
The 22.6\% concentration in Hate \& Harassment (compared to a 6.9\% dataset average, a 3.5$\times$ difference) reflects the prevalence of Russian and Ukrainian identity-based profanity systems, collectively known as \emph{mat}, which include some of the strongest and most targeted slurs in the Slavic language family.
Beyond Cyrillic, the dataset contains content in Polish and Czech (0.30\% of training data), Turkish (0.31\%), and Hungarian (0.30\%).
In the test set, 389 messages (7.2\%) contain Cyrillic script, 22 contain Turkish characters, and 17 contain Polish/Czech characters.

\paragraph{Why XLM-RoBERTa is insufficient.}
A natural question is why the LIM's multilingual lexicon is needed given that M2 (XLM-RoBERTa) is pre-trained on 100 languages including Russian, Ukrainian, Polish, and Turkish.
The answer lies in the distinction between \emph{general-vocabulary} multilingual competence and \emph{domain-specific} profanity detection.
XLM-RoBERTa's pre-training corpus (CommonCrawl) contains formal and semi-formal text, but underrepresents the specific register of gaming chat profanity: context-dependent slurs that are used as identity-based attacks in one context and as general frustration in another, obfuscated forms of profanity, and compound insults that combine multiple languages within a single utterance.
Our validation analysis confirmed this empirically: we identified tokens where $P(\text{toxic} \mid w) = 1 - P(\text{Non-toxic} \mid w) \geq 0.80$ in the training data, yet \emph{both} M1 (DeBERTa) and M2 (XLM-RoBERTa) predicted Non-toxic on the validation set.
The LIM's multilingual lexicon targets precisely these residual blind spots, not as a replacement for XLM-RoBERTa's multilingual capacity, but as a domain-specific complement to it.

\end{document}